\documentclass[conference]{IEEEtran}
\IEEEoverridecommandlockouts
\usepackage{cite}
\usepackage{amsmath,amssymb,amsfonts}
\usepackage{algorithmic}
\usepackage{graphicx}
\usepackage{textcomp}
\usepackage{xcolor}
\def\BibTeX{{\rm B\kern-.05em{\sc i\kern-.025em b}\kern-.08em
    T\kern-.1667em\lower.7ex\hbox{E}\kern-.125emX}}

\usepackage{bm}
\usepackage{graphicx}
\usepackage{amsmath}
\usepackage{multirow}
\usepackage{xspace}
\usepackage{xcolor}
\usepackage{enumitem}
\usepackage{placeins}
\usepackage{booktabs}
\usepackage{hyperref}
\usepackage{cleveref}
  \crefname{Figure}{Fig.}{Fig.}
  
\usepackage{subcaption}
\usepackage[utf8]{inputenc}
\usepackage{pgfplots}
\DeclareUnicodeCharacter{2212}{−}
\usepgfplotslibrary{groupplots,dateplot}
\usetikzlibrary{patterns,shapes.arrows}
\pgfplotsset{compat=newest}

\newcommand{\ours}{{ELTransformer}\xspace}

\begin{document}

\title{Efficient Localness Transformer for Smart Sensor-Based Energy Disaggregation 
}

\author{
  \IEEEauthorblockN{Zhenrui Yue, Huimin Zeng, Ziyi Kou, Lanyu Shang, Dong Wang}
  \IEEEauthorblockA{School of Information Sciences \\
  University of Illinois Urbana-Champaign, Champaign, IL, USA \\
  \{zhenrui3, huiminz3, ziyikou2, lshang3, dwang24\}@illinois.edu}
 }

\maketitle

\begin{abstract}
Modern smart sensor-based energy management systems leverage non-intrusive load monitoring (NILM) to predict and optimize appliance load distribution in real-time. NILM, or energy disaggregation, refers to the decomposition of electricity usage conditioned on the aggregated power signals (i.e., smart sensor on the main channel). Based on real-time appliance power prediction using sensory technology, energy disaggregation has great potential to increase electricity efficiency and reduce energy expenditure.
With the introduction of transformer models, NILM has achieved significant improvements in predicting device power readings. Nevertheless, transformers are less efficient due to $\bm{O(l^{2})}$ complexity w.r.t. sequence length $\bm{l}$. Moreover, transformers can fail to capture local signal patterns in sequence-to-point settings due to the lack of inductive bias in local context. In this work, we propose an efficient localness transformer for non-intrusive load monitoring (\ours). Specifically, we leverage normalization functions and switch the order of matrix multiplication to approximate self-attention and reduce computational complexity. Additionally, we introduce localness modeling with sparse local attention heads and relative position encodings to enhance the model capacity in extracting short-term local patterns. To the best of our knowledge, \ours is the first NILM model that addresses computational complexity and localness modeling in NILM. With extensive experiments and quantitative analyses, we demonstrate the efficiency and effectiveness of the the proposed \ours with considerable improvements compared to state-of-the-art baselines.
\end{abstract}


\section{Introduction}
\label{sec:intro}

\begin{figure}[t]
  \centering
  \includegraphics[width=0.8\linewidth]{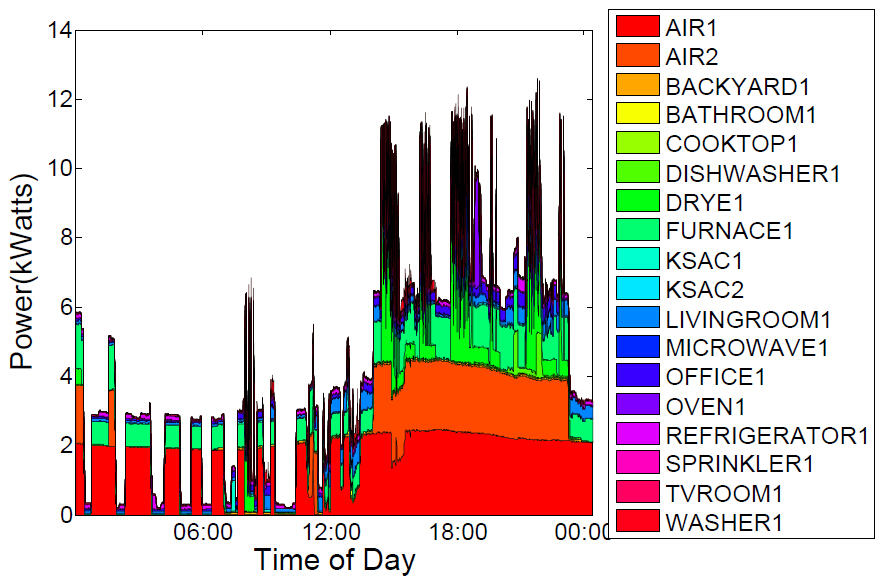}
  \caption{Overview for non-intrusive load monitoring. The aggregated power is decomposed into electricity of various appliances (e.g., washer, oven etc.). Image adopted from \href{http://yeqing.gitlab.io/www/projects.html}{web}.}
  \label{fig:intro}
\end{figure}

Smart energy management systems utilize sensory data to optimize electricity usage for efficient residential energy consumption~\cite{gopinath2020energy}. To better understand household electricity consumption with limited sensory data, non-intrusive load monitoring (NILM, or energy disaggregation) has been proposed to decompose aggregated electricity into individual appliances~\cite{hart1992nonintrusive}. This process is depicted in \Cref{fig:intro}. With smart metering technologies, NILM algorithms predict electricity usage of individual appliances in real-time by monitoring the aggregated consumption in a household~\cite{nardello2017low}. Such sensor-based power analysis provides feedback on electricity usage, and therefore supports sustainable energy consumption~\cite{fischer2008feedback}. 

Machine learning methods have been proposed to improve the prediction accuracy in NILM. Recently, neural networks have been proposed for energy disaggregation and achieved significant improvements compared to traditional machine learning methods. Common neural networks for NILM include long short-term memory (LSTM), denoising autoencoder (DAE) and convolutional neural networks (CNN)~\cite{kolter2012approximate, kelly2015neural, zhang2018sequence}. More recently, transformer models with self-attention mechanism have achieved state-of-the-art performance in multiple natural language processing (NLP) tasks~\cite{vaswani2017attention}. Inspired by self-attention, transformer-based models are designed to improve NILM performance over long input sequences~\cite{yue2020bert4nilm, ccavdar2021efficient}.

Although transformers are light-weight compared to CNN models, an existing challenge for transformer-based NILM methods is the inefficiency of self-attention mechanism~\cite{yue2020bert4nilm}. Standard self-attention utilizes matrix multiplication to compute attention scores. 
For each element, transformer computes corresponding scores with elements across the entire sequence length. This takes $O(l^{2})$ in both time and space for input sequence with length $l$~\cite{vaswani2017attention}. Additionally, transformer models are first designed to capture long-term global context from long sequences. In this sense, they often fail to fully exploit short-term and local signal patterns, which can result in performance deterioration on multi-status appliances~\cite{shin2019subtask, he2021locker}. As a result, current transformer models for NILM only achieve marginal performance gains compared to previous methods\cite{yue2020bert4nilm, ccavdar2021efficient}.

In this paper, we propose an efficient localness transformer model for non-intrusive load monitoring (\ours). In particular, we reduce the computational complexity of transformer models by adapting the order of matrix multiplication and modifying the original self-attention softmax function.
The approximated self-attention significantly reduces computational complexity~\cite{shen2021efficient}. Additionally, we introduce localness modeling with sparse local attention and relative positional encodings, in which sliding windows are adopted to efficiently perform local self-attention and exploit short-term input behavior. To the best of our knowledge, the proposed \ours is the first method that addresses computational complexity and localness modeling in NILM. We demonstrate the efficiency and effectiveness of the proposed \ours with considerable improvements in extensive experiments.


The main contributions of this paper can be summarized as follows:
\begin{enumerate}
\item We propose \ours for non-intrusive monitoring. \ours combines efficient self-attention and local constraints for improved NILM performance.
\item With linear self-attention and light-weight model size, \ours demonstrates significantly improved efficiency in both space and time.
\item We design a localness modeling strategy with local attention heads and relative positional encodings to improve the understanding of local patterns.
\item We demonstrate the efficiency and effectiveness of \ours, where \ours outperforms state-of-the-art baselines on real-world NILM datasets.
\end{enumerate}

\section{Related work}
\label{sec:related_work}

\subsection{Smart Sensing for Energy}
With the advancements in artificial intelligence and internet of things, smart sensors provide increasing potential for sustainable energy management~\cite{gopinath2020energy,zhang2019edgebatch}. For instance, such sensors can be used for detecting appliance status and predicting power consumption values to improve energy efficiency and optimize load distribution~\cite{fischer2008feedback}. Early smart meters and sensor systems are designed to measure appliance electricity consumption in households or buildings~\cite{benyoucef2010smart, wang2012energy}. Intelligent sensor systems that include both hardware and software design are proposed for energy disaggregation and event-based analysis~\cite{nardello2017low,rashid2019sead}. Recently, sensor-based energy management is proposed to reduce overall consumption and optimize operation efficiency in the development of smart grids and smart cities~\cite{gopinath2020energy,zhang2018real}.

\subsection{Energy Management via NILM}
Energy management is possible via smart sensing with increasing computing power in SoCs and efficient NILM algorithms within such systems~\cite{nardello2017low}. Most NILM methods have been proposed in 3 different settings: (1)~Sequence-to-sequence (Seq2seq), where NILM model performs regression upon a sequence and output a sequence of appliance power within the same time frame~\cite{kelly2015neural}; (2)~Sequence-to-subsequence (Seq2subseq), in which output signals are of a shorter sequence length~\cite{pan2020sequence} and (3)~Sequence-to-point (Seq2point), where the model learns to predict a certain point (e.g., the midpoint) from input sequences~\cite{zhang2018sequence}. Nevertheless, NILM algorithms in energy management systems rely on heuristics or expensive neural networks, resulting in less accurate predictions or demands for extensive computational resources. As such, we propose \ours, an efficient transformer network for seq2point energy disaggregation, which can be integrated in smart sensor systems with limited computing power.

\subsection{Neural Networks for NILM}
Various neural networks have been proposed for energy disaggregation, we summarize them according to the adopted network architecture: (1)~Autoencoder: Autoencoder models attempt to rebuild the target appliance signals from the aggregated main readings. Such methods learn a low dimensional feature representation and perform regression on the target appliance~\cite{kelly2015neural}; (2)~Recurrent neural network: LSTM is first proposed to handle long signal sequences~\cite{kelly2015neural}. Further research on LSTM and GRU models improves real-time inference and generalization by imposing regularization~\cite{rafiq2018regularized}; (3)~Convolutional neural network (CNN): Seq2point proposes a CNN model to predict power at the sequence midpoint with convolutional layers and linear transformations~\cite{zhang2018sequence}. CNN-based generative adversarial network avoids designing an objective function by adopting an additional discriminator~\cite{pan2020sequence}. Existing neural networks are extremely large and complex in design, which render these techniques widely inaccessible for efficient energy disaggregation. In contrast to previous efforts, the proposed \ours has less than 2M parameters and can efficiently perform energy disaggregation in linear time. 




\subsection{Transformer Models in NILM}
Transformer models with attention mechanism have been proposed as language models to process long input sequences~\cite{vaswani2017attention}. Transformer models for NILM adopt convolutional layers to extract features from the input sequence. The following transformer layers use self-attention to process the sequential input. In particular, self-attention matches every input element with the entire input sequence to compute attention scores.
The output sequence is then generated by a decovolutional layer and linear transformation~\cite{ccavdar2021efficient}.
Nevertheless, for sequence with length $l$, 
self-attention operation yields $O(l^{2})$ complexity in both space and time. This complexity leads to significantly slower training and inference time upon deployment.
For the seq2point setting in this paper, another drawback of the transformer model is the lack of localness modeling. Although self-attention is designed to capture long-term semantics from input sequences, they often fail to capture short-term and local signal patterns~\cite{he2021locker}. In energy disaggregation, lacking local dependency can lead to mismatches and performance drops for multi-status appliances~\cite{shin2019subtask}. To address computational complexity and localness modeling, the proposed \ours leverages efficient global attention with $O(l)$. Additional local attention is designed to introduce local inductive bias and exploit short-term local patterns for improved multi-status appliance prediction.


Unlike previous transformer models in NILM, we propose an efficient localness transformer model for non-intrusive load monitoring (\ours). We reduce the computational costs of training a transformer model and address localness modeling in the seq2point setting for NILM. Specifically, we leverage normalization functions and switch the order of self-attention multiplication to approximate global attention mechanism.
Additionally, local attention heads and relative positional encodings are introduced to capture short-term local signal patterns. 
To the best of our knowledge, the proposed \ours is the first method that addresses efficiency and localness modeling for NILM algorithms.
\section{Methodology}
\label{sec:method}

\begin{figure}
  \centering
  \includegraphics[width=0.6\linewidth]{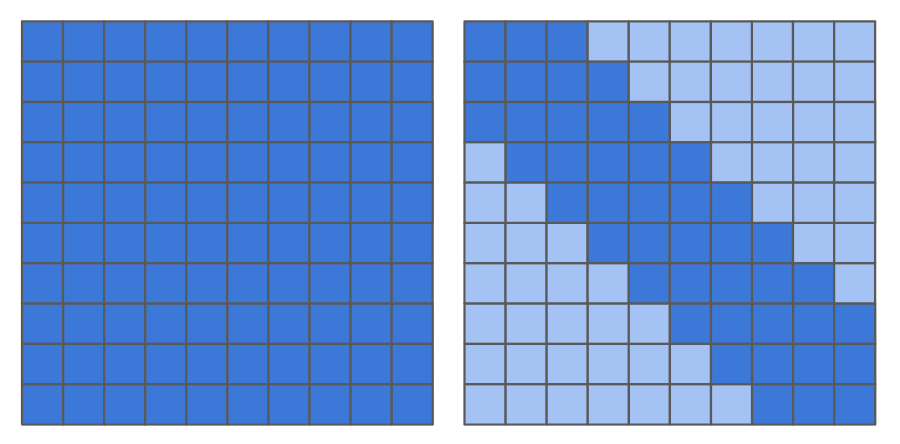}
  \caption{Global attention vs. local attention, dark blue indicates positions that can be attended. For local attention, only elements from a small neighborhood can be attended to compute attention scores.}
  \label{fig:local_attn}
\end{figure}

\subsection{Problem Definition}
Non-intrusive load monitoring can be formulated as decomposition of the sum of a household's energy consumption into the consumption of individual appliances. Input is given by a sequence of aggregated energy consumption $\bm{x}$ (Note that the length of $\bm{x}$ should be odd for midpoint prediction), while $\bm{y}_{i}$ indicates the i-th appliance usage readings, therefore:
\begin{equation}
  \sum_{i=1}^{N} \bm{y}_i = \bm{x} + \bm{e},\\
\end{equation}
where $N$ is the sum of all appliances and $\bm{e}$ stands for noise. In the seq2point setting, the model predicts power value for a certain appliance at the midpoint of $\bm{x}$. Therefore, we replace $\bm{y}$ and denote the scalar output with $y$ in the following formulation, namely $y = \bm{y}_{(|\bm{y}|+1)/2}$. Given NILM model $\bm{f}$, ideally, we have $y = \bm{f}(\bm{x})$.

For training NILM models, we want $\bm{f}$ to predict appliance usage $y$ as accurate as possible. In other words, we 
minimize the mean squared error (MSE) between model output and ground truth energy consumption $y$. Therefore, we can formulate the training as an an optimization problem w.r.t. input data and $\bm{f}$ as follows:
\begin{equation}
    \bm{f}^{*} = \arg \min_{\substack{\bm{f}}} \sum_{i=1}^{|\bm{X}|} \mathcal{L}_{\mathrm{mse}}(\bm{f}(\bm{x}_{i}), y_{i}),
\end{equation}
where $\bm{x}_{i}$ indicates the i-th input sequence in dataset $\bm{X}$, and $y_{i}$ refers to the ground truth power value at midpoint of $\bm{x}_{i}$.

\subsection{Linear Attention}
Self-attention can be formulated with input-based matrices $\bm{Q}$, $\bm{K}$ and $\bm{V}$:
\begin{equation}
  \mathrm{Attn}(\bm{Q}, \bm{K}, \bm{V}) = \sigma (\frac{\bm{Q}\bm{K}^{T}}{\sqrt{d}})\bm{V},
  \label{eq:attn}
\end{equation}
where $\sigma$ represents the softmax function and $d$ refers to the feature dimensionality. 
The similarity between the i-th query $\bm{q}_{i}$ and the j-th key $\bm{k}_{j}$ can be measured with $\bm{q}_{i} \bm{k}_{j}$, this is then scaled with softmax function to compute attention scores. However, since every query in $\bm{Q}$ is matched with every key in $\bm{K}$, this corresponds to $l^{2}$ times of matching and result in complexity $O(l^{2})$ w.r.t. input length $l$~\cite{vaswani2017attention}. 

To improve the efficiency of transformer-based model, we adopt efficient attention to accelerate the computation of self-attention~\cite{shen2021efficient}. Specifically, we introduce normalization functions $\rho$ and switch the order of self-attention:
\begin{equation}
  \mathrm{Attn}(Q, K, V) = \rho_{q}(\bm{Q})(\rho_{k}(\bm{K})^{T} \bm{V}),
  \label{eq:effi_attn}
\end{equation}
where $\rho_{q}$ and $\rho_{k}$ are query and key normalization functions. The $\rho$ functions normalize $\bm{Q}$ and $\bm{K}$ separately to approximate the original softmax function in \Cref{eq:attn}, they can be formulated as follows:
\begin{equation}
  \begin{aligned}
    \rho_{q}(\bm{Q}) &= \sigma_{row}(\bm{Q}), \\
    \rho_{k}(\bm{K}) &= \sigma_{col}(\bm{K}), \\
  \label{eq:ours_loss}
  \end{aligned}
\end{equation}
$\sigma$ represents the softmax function. $\sigma_{row}$ performs softmax w.r.t. the row dimension in $\bm{Q}$, while $\sigma_{col}$ computes softmax in the column dimension.

The linear attention module is mathematically equivalent to the original self-attention, we only switch the order from $(\bm{Q}\bm{K}^{T})\bm{V}$ to $\bm{Q}(\bm{K}^{T}\bm{V})$. For a fixed hidden dimension $d$ (i.e., matrices are of dimension $l \times d$), it approximates the original self-attention and reduces the quadratic computational complexity to linear complexity. With increasing input lengths, the linear attention module can significantly reduce computational costs for training and inference. In our implementation, we split the hidden dimension and compute self-attention multiple times (i.e., multi-head attention), their output matrices are concatenated to build the attentive output.

\begin{figure*}
  \centering
  \includegraphics[width=\linewidth]{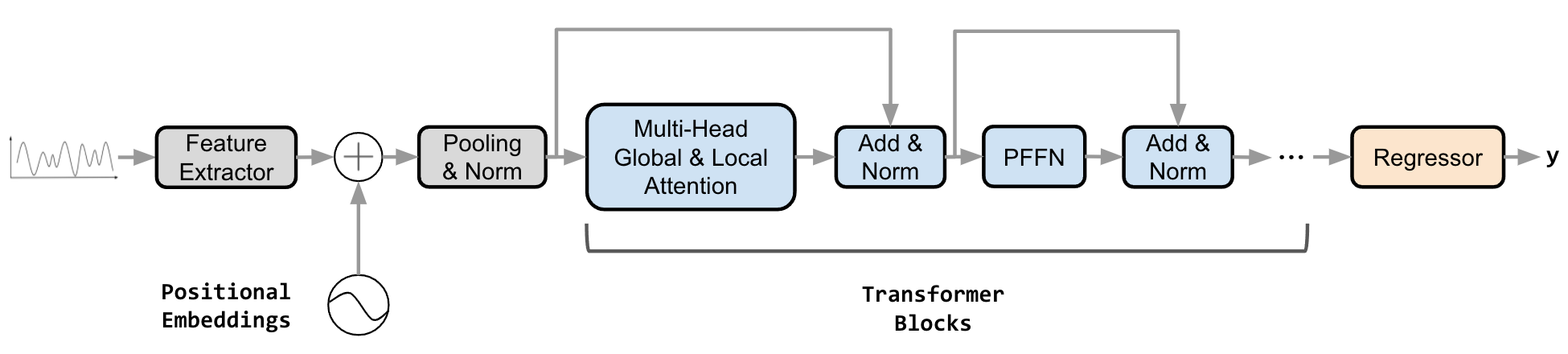}
  \caption{Architecture of \ours.}
  \label{fig:model}
\end{figure*}

\subsection{Localness Modeling}
While self-attention mechanism is designed to capture global context from input sequence, it does not perform well in capturing local signal patterns. For instance, NILM methods often fail to detect status changes on multi-status appliances (e.g., standby/on in washing machine). This can be traced back to the lack of inductive bias in local context, as self-attention is designed to be a global operator~\cite{he2021locker}. To alleviate this issue, local attention heads are introduced.

Different from previous methods, we leverage local attention heads to enhance localness modeling~\cite{child2019generating}. Unlike global attention, local attention splits the input sequence into smaller data chunks of a fixed local window size $l_{\mathrm{win}}$. In each window, local attention can attend elements from itself and its neighboring windows instead of the whole sequence, see \Cref{fig:local_attn}. Specifically for $\bm{Q}_{\mathrm{loc}}$, $\bm{K}_{\mathrm{loc}}$ and $\bm{V}_{\mathrm{loc}}$ in local attention heads, we first perform padding (to ensure that the input can be evenly divided by the local window size) and split the entire input sequences into local windows of size $l_{\mathrm{win}}$. Then, we expand $\bm{K}_{\mathrm{loc}}$ and $\bm{V}_{\mathrm{loc}}$ to include elements from neighboring windows, followed by the computation of self-attention scores in each local window with standard self attention, i.e.:
\begin{equation}
  \mathrm{Attn_{loc}}(\bm{Q}_{\mathrm{loc}}, \bm{K}_{\mathrm{loc}}, \bm{V}_{\mathrm{loc}}) = \sigma (\frac{\bm{Q}_{\mathrm{loc}}\bm{K}_{\mathrm{loc}}^{T}}{\sqrt{d_{\mathrm{loc}}}})\bm{V}_{\mathrm{loc}},
  \label{eq:local_attn}
\end{equation}
where $d_{\mathrm{loc}}$ is the dimension of local attention head. $\bm{Q}_{\mathrm{loc}}$ is of dimension $l_{\mathrm{win}} \times d_{\mathrm{loc}}$, $\bm{K}_{\mathrm{loc}}$ and $\bm{V}_{\mathrm{loc}}$ are of dimension $3l_{\mathrm{win}} \times d_{\mathrm{loc}}$. The sparsity in the attention matrix significantly accelerates the computation of attention scores. Additionally, this reduces the complexity of self attention. For fixed hidden dimension $d_{\mathrm{loc}}$ and window size $l_{\mathrm{win}}$, the local attention has a linear computational complexity.

In addition to local attention heads, we incorporate local constraints in the regressor with relative positional embeddings to improve localness modeling. In particular, the learnable positional encodings are symmetric w.r.t. the midpoint of the sequence. We add positional encodings to the transformer output before performing regression. With limited additional parameter, the local attention heads and relative positional encodings introduce additional inductive bias to the proposed \ours for capturing short-term signal behavior.

\subsection{Model Architecture}
The proposed \ours model is depicted in \Cref{fig:model}. Unlike the previous transformer models, we adopt both global and local attention heads to address the lack of localness modeling in NILM methods. Moreover, we adopt efficient self-attention to achieve linear computational complexity, which greatly reduces the training and inference costs. We also adopt relative positional encodings in the regressor to introduce additional inductive bias for seq2point NILM. We elaborate the specifics of the proposed \ours in the following.

Similar to~\cite{yue2020bert4nilm}, we first extract features from the one-dimensional input sequence by adopting two 1d-convolutional layers. We concatenate the extracted features from convolution layers. In the feature extractor, we use learnable embeddings to indicate input positions and convert positional information to vectors of the same dimension. The positional embeddings are added to extracted features, followed by $l^{2}$ pooling over the length dimension of the input sequence. Pooling on the one hand reduces the sequence length, it also accelerates forward feeding in attention blocks. A linear transformation is followed. The above feature extractor can be formulated as:
\begin{equation}
  \mathrm{E}(\bm{x}) = \mathrm{Pool}(\mathrm{Cat}(\mathrm{Conv_{1}}(\bm{x}), \mathrm{Conv_{2}}(\bm{x})) + \bm{E}_{pos})\bm{W} + \bm{b},
\end{equation}
where Pool stands for pooling, Cat stands for concatenation and Conv stands for convolutional operators. $\bm{W}$ and $\bm{b}$ are learnable parameters for linear transformation, $\bm{E}_{pos}$ represents the positional embeddings. In the following, we denote the extracted high-dimensional input data with $\bm{X}$.

The following layers are built based on transformer blocks. In the proposed \ours, the most important operations for transformer blocks are global and local self-attention operation (or scaled dot-product attention). Global attention head can be formulated with the efficient linear self-attention, as in \Cref{eq:effi_attn}. For local attention within each local window, full self-attention can be performed as in \Cref{eq:local_attn}. We adopt multi-head attention to build global attention heads and local attention heads in parallel. In each attention head, the matrices $\bm{Q}$ (Query), $\bm{K}$ (Key) and $\bm{V}$ (Value) can be obtained by 3 different linear transformation based on the input matrix $\bm{X}$~\cite{vaswani2017attention}. The attentive output matrices from global and local heads are concatenated. 

In addition to self-attention, position-wise feed forward networks (PFFN) are applied, PFFNs are defined as:
\begin{equation}
  \mathrm{PFFN}(X) = \mathrm{GELU}(\bm{X}\bm{W}_{1} + \bm{b}_{1})\bm{W}_{2} + \bm{b}_{2}),
\end{equation}
In PFFN, the inner-layer dimensionality is 4 times the size of input and output dimensionality, we perform activation with GELU activation function~\cite{hendrycks2016gaussian}. Note that after self-attention operation and PFFNs, residual connections and layer normalization (LayerNorm) are applied to improve training (see blue blocks in \Cref{fig:model}). For a module (i.e., Attention or PFFN), this can be formulated as:
\begin{equation}
  \mathrm{LayerNorm}(\bm{X} + \mathrm{Module}(\bm{X})).
\end{equation}

In the final regressor layers, since we adopt the seq2point setting, relative position embeddings are applied and added to the transformer output. The relative position embeddings are learnable embeddings with symmetric values along the column dimension (i.e., embeddings are identical when elements have same distance to the sequence midpoint). We then perform layer normalization and feed data to a 2-layer multilayer perceptron (MLP) for the regression value,
\begin{equation}
  \mathrm{Out}(\bm{X}) = \mathrm{ReLU}(\mathrm{LayerNorm}(\bm{X} + \bm{E}_{rel})\bm{W}_{1} + \bm{b}_{1})\bm{W}_{2} + b_{2},
\end{equation}
with $\bm{E}_{rel}$ and ReLU representing the relative position embeddings and ReLU activation function respectively.

\section{Evaluation}
\label{sec:experiment}

\subsection{Datasets}
We select two frequently used energy disaggregation datasets to validate the proposed \ours:
\begin{itemize}
  \item UK-DALE~\cite{kelly2015uk}: UK-DALE measures the domestic appliance-level and whole-house electricity demand of five homes in the UK. 
  In our experiments, we adopt low-frequency data. The target appliances are: dishwasher, fridge, kettle, microwave and washing machine (washer).
  \item REDD~\cite{kolter2011redd}: REDD is one of the first datasets with freely available residential power usage data from six houses in US. 
  We experiment on the resampled low-frequency readings and experiment on the same appliances as in UK-DALE except kettle.
\end{itemize}

To evaluate the model generalizability, we selected different households for training and evaluation. Specifically, we reserve 1 unseen household in each dataset only for evaluation, see \Cref{tab:train-test-distribution}~\cite{kelly2015neural}. For UK-DALE, we select active power for both aggregated mains and appliance channels. In our experiments, we exclude house 3 and 4, as active power is not available in the mains. In REDD, only apparent power is available for mains. As such, we use apparent power for mains and active power for appliance power readings. For training, we use data from the last 2 recorded months of training houses in UK-DALE and the last recorded month of training houses in REDD~\cite{shin2019subtask}. For the separate testing households, we use data from the last recorded month.

\begin{table}[t]
  \centering
  \caption{Training and evaluation data distribution. Date from the selected test households is only used for evaluation.}
  \begin{tabular}{lccc}
    \toprule
    Name & Train \& Validation & Test \\
    \midrule
    UK-DALE & 1, 5 & 2 \\
    REDD & 2, 3, 4, 5, 6 & 1 \\
    \bottomrule
  \end{tabular}
  \label{tab:train-test-distribution}
\end{table}

\subsection{Baselines}
For baseline methods, we select 3 state-of-the-arts methods for the seq2point setting:
\begin{itemize}
  \item Seq2point~\cite{zhang2018sequence}: Seq2point is a CNN-based neural network for energy disaggregation. Seq2point first proposes sequence-to-point learning in NILM with a simple network of convolutional layers and an MLP regressor. 
  \item SGNet~\cite{shin2019subtask}: Subtask gated network (SGNet) comprises of 2 similar seq2point CNN networks. SGNet multiplies the main network’s power prediction with the subtask’s status prediction. With subtask network, SGNet achieves improved performance on multi-status appliance. 
  \item BERT4NILM~\cite{yue2020bert4nilm}: BERT4NILM is the first transformer network applied in NILM. BERT4NILM leverages convolutional layers and bidirectional self-attention with multiple attention heads to predict appliance power. BERT4NILM is trained with masked training process, where input elements are randomly masked.
\end{itemize}

For SGNet, we modify the last layer following~\cite{zhang2018sequence} and adapt it for seq2point, we train SGNet with both MSE loss and binary cross entropy (BCE) loss~\cite{shin2019subtask}. The original BERT4NILM produces full sequences as output. In our experiments, we take output from the first position (comparable to [CLS] token in NLP) and optimize the model with MSE loss to adapt it to the seq2point setting. 

\subsection{Implementation}
\subsubsection{Preprocessing} The original data from both REDD and UK-DALE is sampled at least every 6 seconds on both aggregated channel and individual appliances. Following~\cite{kelly2015neural}, we align the timestamps and resample every 6 seconds with the average value method to alleviate the influences of outlier samples. We drop timestamps with NaN values. For power readings, we normalize the input data by subtracting the mean and then dividing it by the standard deviation of the corresponding appliance. Mean and standard deviation values of all channels from training data are preserved for evaluation. To generate seq2point data samples, a sliding window approach is adopted to generate input samples with fixed length of 599, ground truth appliance value from the corresponding midpoint is used as label~\cite{zhang2018sequence}. Such individual samples would be used to form mini-batches of size 256 in training. Data preprocessing is implemented with NILMTK~\cite{batra2014nilmtk}.

\subsubsection{Training and evaluation}
From the training data, we reserve 20\% for validation and the rest can be used for training. \ours is initiated with 256 hidden dimensionality and 4 attention heads using local window size of 20. Baseline models are initiated with default hyperparameter. All models except SGNet are trained with MSE loss, SGNet is trained with both MSE loss and BCE loss. Training is performed with the learning rate of 1e-4 without learning rate decay. The Adam optimizer is adopted for optimization, with betas of 0.9 and 0.999 and zero weight decay. Validation is performed after each epoch and we save the model with the best validation loss. Early stopping is called when validation loss is not improved after 5 consecutive epochs. 

\begin{table}[t]
  \centering
  \caption{On thresholds of different appliances.}
  \begin{tabular}{lccccc}
    \toprule
    Name & Dishwasher & Fridge & Kettle & Microwave & Washer \\
    \midrule
    Thres. & 10W & 50W & 2000W & 200W & 20W \\
    \bottomrule
  \end{tabular}
  \label{tab:on-thresholds}
\end{table}

To evaluate the proposed \ours model, we adopt 3 common metrics used in NILM, mean absolute error (MAE), F1 score and Matthews correlation coefficient (MCC). 
In evaluation, we first evaluate MAE with model output. We then compute on/off status by comparing predictions with on threshold of each target appliance, see \Cref{tab:on-thresholds}. Then, true positive (TP), true negative (TN), false positive (FP) and false negative (FN) are summarized to compute F1 and MCC.

\subsection{Quantitative Results}
\begin{table}[t]
  \centering
  \caption{Main results on UK-DALE.}
  \label{tab:uk-dale-results}
  \begin{tabular}{lcccc}
  \toprule
  \textbf{Appliance}          & \textbf{Model} & \textbf{MAE} $\downarrow$ & \textbf{F1} $\uparrow$ & \textbf{MCC} $\uparrow$ \\ \midrule
  \multicolumn{5}{c}{(I) Average Performance}                                                            \\ \midrule
  \multirow{4}{*}{Average}    & Seq2Point      & \underline{14.33} & 0.63             & \underline{0.65} \\
                              & SGNet          & 15.37             & \underline{0.64} & 0.64             \\
                              & BERT4NILM      & 18.59             & 0.58             & 0.59             \\
                              & \ours (Ours)   & \textbf{12.92}    & \textbf{0.65}    & \textbf{0.66}    \\ \midrule
  \multicolumn{5}{c}{(II) Appliance-specific Performance}                                                \\ \midrule
  \multirow{4}{*}{Dishwasher} & Seq2Point      & 24.43             & 0.70             & 0.71             \\
                              & SGNet          & \underline{23.66} & \underline{0.77} & \underline{0.76} \\
                              & BERT4NILM      & 22.03             & 0.74             & 0.74             \\
                              & \ours (Ours)   & \textbf{17.89}    & \textbf{0.79}    & \textbf{0.78} \\ \midrule
  \multirow{4}{*}{Fridge}     & Seq2Point      & \underline{12.97} & \textbf{0.92}    & \textbf{0.88}    \\
                              & SGNet          & 16.59             & 0.89             & 0.84             \\
                              & BERT4NILM      & 15.60             & 0.86             & 0.79             \\
                              & \ours (Ours)   & \textbf{12.27}    & \underline{0.91} & \underline{0.86} \\ \midrule
  \multirow{4}{*}{Kettle}     & Seq2Point      & \underline{7.05}  & 0.88             & 0.89             \\
                              & SGNet          & 7.34              & \underline{0.89} & \underline{0.90} \\
                              & BERT4NILM      & 7.98              & 0.89             & 0.90             \\
                              & \ours (Ours)   & \textbf{6.95}     & \textbf{0.95}    & \textbf{0.95}    \\ \midrule
  \multirow{4}{*}{Microwave}  & Seq2Point      & 9.43              & 0.26             & 0.30             \\
                              & SGNet          & \underline{9.15}  & \underline{0.31} & \underline{0.34} \\
                              & BERT4NILM      & 19.96             & 0.23             & 0.30             \\
                              & \ours (Ours)   & \textbf{8.78}     & \textbf{0.39}    & \textbf{0.42}    \\ \midrule
  \multirow{4}{*}{Washer}     & Seq2Point      & \textbf{17.76}    & \textbf{0.43}    & \textbf{0.50}    \\
                              & SGNet          & 20.15             & \underline{0.37} & \underline{0.38} \\
                              & BERT4NILM      & 27.38             & 0.18             & 0.23             \\
                              & \ours (Ours)   & \underline{18.74} & 0.21             & 0.28             \\ \bottomrule
\end{tabular}
\end{table}

\begin{table}[t]
  \centering
  \caption{Main results on REDD.}
  \label{tab:redd-results}
  \begin{tabular}{lcccc}
  \toprule
  \textbf{Appliance}          & \textbf{Model} & \textbf{MAE} $\downarrow$ & \textbf{F1} $\uparrow$ & \textbf{MCC} $\uparrow$ \\ \midrule
  \multicolumn{5}{c}{(I) Average Performance}                                                            \\ \midrule
  \multirow{4}{*}{Average}    & Seq2Point      & 37.13             & \underline{0.39} & 0.33             \\
                              & SGNet          & \underline{36.70} & 0.38             & \underline{0.37} \\
                              & BERT4NILM      & 39.02             & 0.37             & 0.34             \\
                              & \ours (Ours)   & \textbf{34.71}    & \textbf{0.41}    & \textbf{0.40}    \\ \midrule
  \multicolumn{5}{c}{(II) Appliance-specific Performance}                                                \\ \midrule
  \multirow{4}{*}{Dishwasher} & Seq2Point      & 22.75             & \underline{0.41} & 0.40             \\
                              & SGNet          & 21.35             & 0.39             & \underline{0.45} \\
                              & BERT4NILM      & \underline{19.85} & 0.40             & 0.40             \\
                              & \ours (Ours)   & \textbf{18.18}    & \textbf{0.44}    & \textbf{0.49}    \\ \midrule
  \multirow{4}{*}{Fridge}     & Seq2Point      & 66.76             & 0.51             & 0.34             \\
                              & SGNet          & \underline{65.75} & 0.50             & 0.35             \\
                              & BERT4NILM      & 67.28             & \underline{0.51} & \underline{0.36} \\
                              & \ours (Ours)   & \textbf{65.52}    & \textbf{0.51}    & \textbf{0.37}    \\ \midrule
  \multirow{4}{*}{Microwave}  & Seq2Point      & 24.14             & \underline{0.27} & 0.26             \\
                              & SGNet          & 19.66             & 0.25             & \underline{0.28} \\
                              & BERT4NILM      & \underline{17.88} & 0.22             & 0.24             \\
                              & \ours (Ours)   & \textbf{15.82}    & \textbf{0.29}    & \textbf{0.34}    \\ \midrule
  \multirow{4}{*}{Washer}     & Seq2Point      & \textbf{34.89}    & 0.35             & 0.34             \\
                              & SGNet          & 40.04             & \underline{0.38} & \textbf{0.41}    \\
                              & BERT4NILM      & 51.10             & 0.36             & 0.35             \\
                              & \ours (Ours)   & \underline{39.35} & \textbf{0.40}    & \underline{0.39} \\ \bottomrule
\end{tabular}
\end{table}


\begin{figure}[h]
  \centering
  \includegraphics[width=0.7\linewidth]{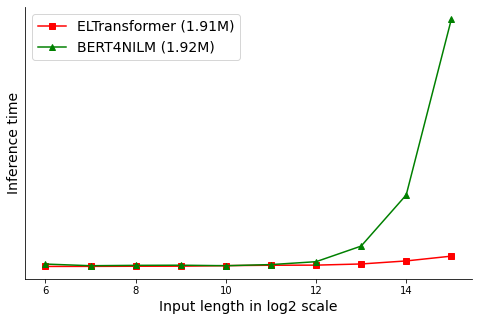}
  \caption{Inference time with different input length.}
  \label{fig:efficiency}
\end{figure}

\begin{figure*}[t]
\centering
    \begin{subfigure}[b]{0.3\textwidth}
      \centering
      \includegraphics[width=\textwidth]{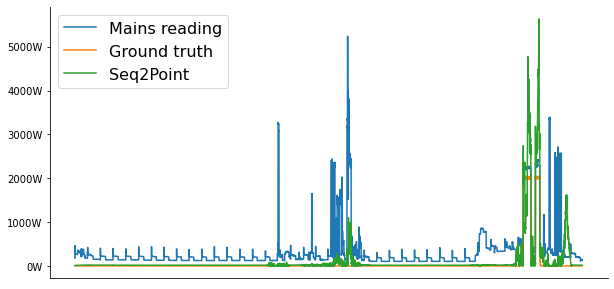}
      \caption{Seq2Point on dishwasher}
	\end{subfigure}
	\begin{subfigure}[b]{0.3\textwidth}
      \centering
      \includegraphics[width=\textwidth]{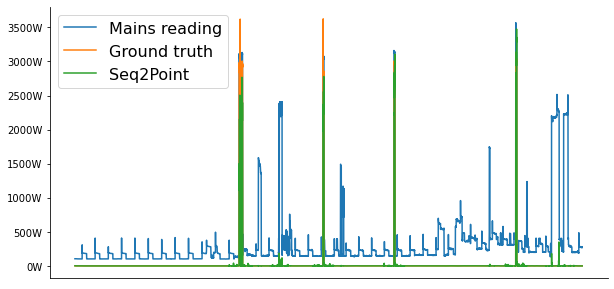}
      \caption{Seq2Point on kettle}
	\end{subfigure}
	\begin{subfigure}[b]{0.3\textwidth}
      \centering
      \includegraphics[width=\textwidth]{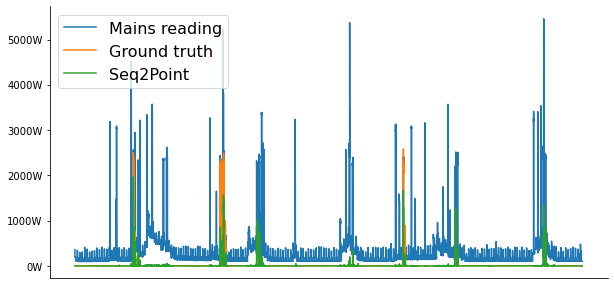}
      \caption{Seq2Point on washer}
	\end{subfigure}
	\begin{subfigure}[b]{0.3\textwidth}
      \centering
      \includegraphics[width=\textwidth]{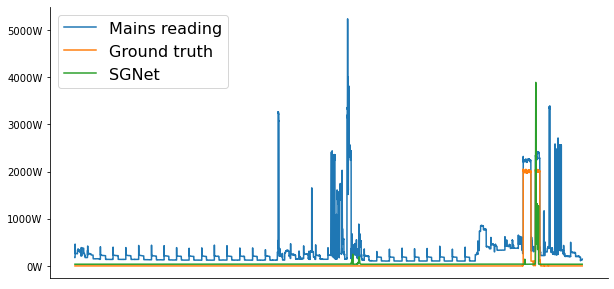}
      \caption{SGNet on dishwasher}
	\end{subfigure}
	\begin{subfigure}[b]{0.3\textwidth}
      \centering
      \includegraphics[width=\textwidth]{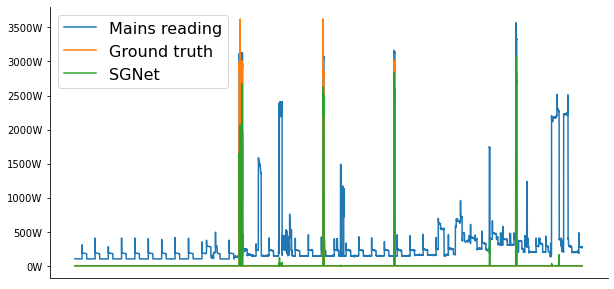}
      \caption{SGNet on kettle}
	\end{subfigure}
	\begin{subfigure}[b]{0.3\textwidth}
      \centering
      \includegraphics[width=\textwidth]{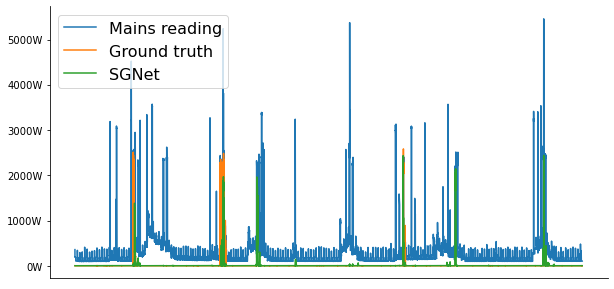}
      \caption{SGNet on washer}
	\end{subfigure}
	\begin{subfigure}[b]{0.3\textwidth}
      \centering
      \includegraphics[width=\textwidth]{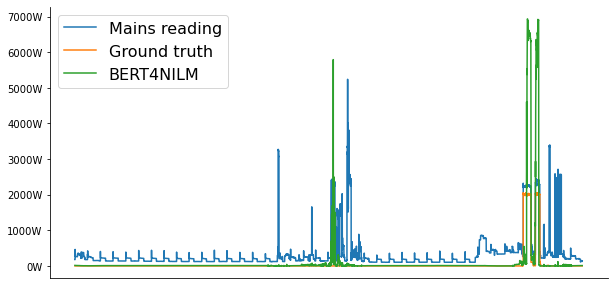}
      \caption{BERT4NILM on dishwasher}
	\end{subfigure}
	\begin{subfigure}[b]{0.3\textwidth}
      \centering
      \includegraphics[width=\textwidth]{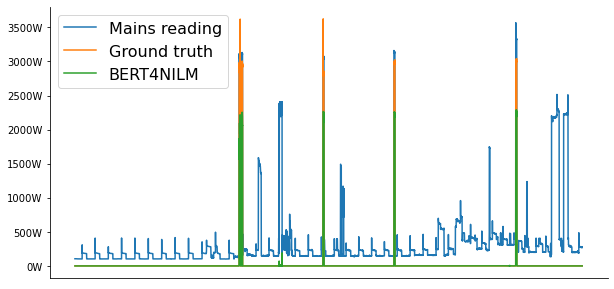}
      \caption{BERT4NILM on kettle}
	\end{subfigure}
	\begin{subfigure}[b]{0.3\textwidth}
      \centering
      \includegraphics[width=\textwidth]{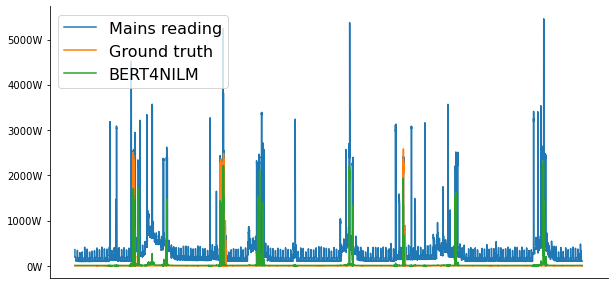}
      \caption{BERT4NILM on washer}
	\end{subfigure}
	\begin{subfigure}[b]{0.3\textwidth}
      \centering
      \includegraphics[width=\textwidth]{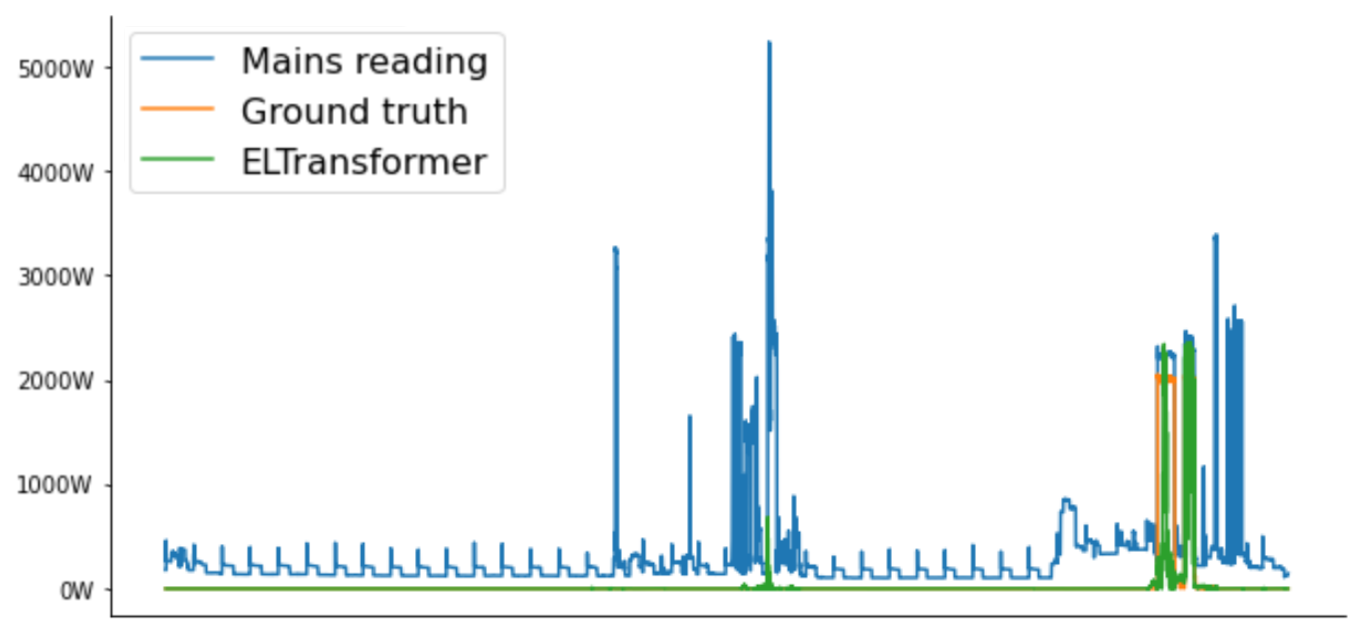}
      \caption{\ours on dishwasher}
	\end{subfigure}
	\begin{subfigure}[b]{0.3\textwidth}
      \centering
      \includegraphics[width=\textwidth]{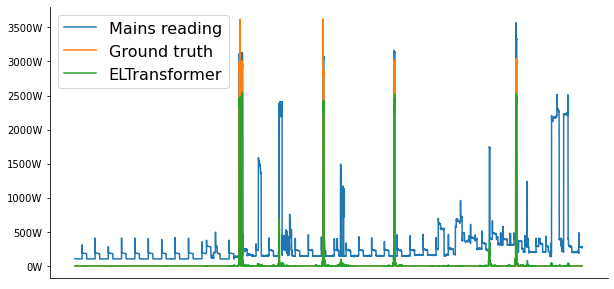}
      \caption{\ours on kettle}
	\end{subfigure}
	\begin{subfigure}[b]{0.3\textwidth}
      \centering
      \includegraphics[width=\textwidth]{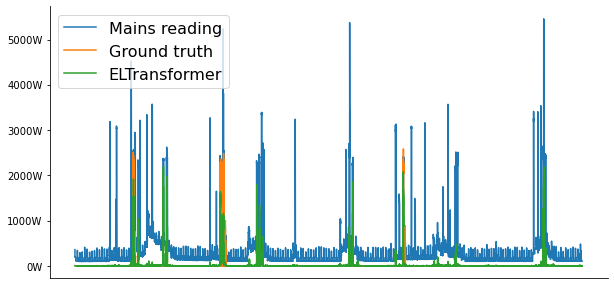}
      \caption{\ours on washer}
	\end{subfigure}
\caption{Qualitative results of baselien methods and \ours on dishwasher, kettle and washing machine.}
\label{fig:qualitative}
\end{figure*}

\Cref{tab:uk-dale-results} and \Cref{tab:redd-results} report the main results on UK-DALE and REDD, each table can be divided into two parts: (1)~Average results and (2)~Appliance-specific performance. Best results are marked in bold, second best results are underlined. From the main results we can make the following observations: (1)~The baseline methods perform generally well and provide a challenge for \ours. For example, Seq2point outperforms \ours on washing machine in UK-Dale; (2)~On average, the proposed \ours performs the best among all methods. Compared to the second best model, \ours achieves 8.2\%, 4.2\% and 4.8\% improvements in MAE, F1 and MCC respectively; (3)~\ours performs consistently well on different appliances, outperforming all baseline methods in MAE results with the exception of washer; (4)~Even with significantly reduced model size (1.91M parameter), \ours delivers similar or better results than CNN models (Seq2point: 30.7M, SGNet: 61.4M, BERT4NILM: 1.92M).

We also study the computational efficiency of the proposed \ours with BERT4NILM, since they have similar architecture and number of parameters. In our experiments, we observe that for fixed input length 599, \ours is 27.8\% faster in training and 32.3\% faster in inference than BERT4NILM. Additionally, we examine the inference efficiency for varying input lengths, see \Cref{fig:efficiency}. We observe that when sequences are shorter than $2^{12}$, \ours performs on average similarly to CNN models ($O(l)$ computational complexity) and outperforms BERT4NILM. For longer sequences, the inference time of BERT4NILM starts growing quadratically, while \ours remains efficient, suggesting that \ours has similar efficiency to CNN while being much more scalable with only 1.91M parameters.


In summary, the main results demonstrate that the proposed localness modeling approach is effective for seq2point NILM. Additionally, the improved efficiency in \ours with linear self-attention leads to significantly reduced computational costs for both training and inference.

\subsection{Qualitative Results}
We present qualitative examples of the predicted power signals using baseline methods and the proposed \ours in \Cref{fig:qualitative}.
We take 3 appliances in UK-DALE and visualize the output signals. Y-axis represents appliance power and x-axis represents time. The selected appliances are dishwasher, kettle and washing machine (washer). For comparison, we add the ground truth and the aggregated mains readings.

From the qualitative results, we can make a few interesting observations. First, we see predicted signals from baseline methods are generally higher and contain more noise on dishwasher. This may be attributed to the lack of local inductive bias in baseline models. Second, the predicted readings from \ours are accurate on the selected appliances, the predictions are more consistent with the ground truth and yield similar trends over time, this indicates that the combination of global and local attention improves prediction on multi-status appliances like dishwasher.

\subsection{Ablation Studies}


\subsubsection{Local Attention Heads}
\begin{table}[h]
  \centering
  \caption[Sensitivity of local attention]{Evaluation performance with different number of local attention heads.}
  \begin{tabular}{ccc}
  \toprule
  \multirow{2}{*}{\textbf{Local head}} & \multicolumn{1}{c}{\textbf{Dishwasher}} & \textbf{Fridge} \\
  & \textbf{MAE} $\downarrow$ / \textbf{F1} $\uparrow$ / \textbf{MCC} $\uparrow$ & \textbf{MAE} $\downarrow$ / \textbf{F1} $\uparrow$ / \textbf{MCC} $\uparrow$ \\ \midrule
  0  & 23.71 / 0.40 / 0.38                                     & 66.76 / 0.51 / 0.36                            \\
  1  & 19.85 / 0.41 / 0.48                                     & 68.27 / 0.51 / 0.36                            \\
  2  & \underline{19.10} / \underline{0.44} / \underline{0.48} & \textbf{65.52} / \textbf{0.51} / \textbf{0.37} \\ 
  3  & 21.71 / 0.38 / 0.45                                     & 68.66 / \underline{0.51} / \underline{0.37}    \\
  4  & \textbf{18.18} / \textbf{0.44} / \textbf{0.49}          & \underline{66.71} / 0.51 / 0.34    \\ \bottomrule
  \end{tabular}
  \label{tab:local-attention}
\end{table}
To study the influence of the number of local attention heads, we initialize 4 attention heads in total and increase the number of local attention heads to study the effectiveness of localness modeling. Then we perform training and evaluate the performance on selected appliances, we adopt dishwasher and fridge in REDD as target appliances. Results are presented in \Cref{tab:local-attention}. We observe that local attention can improve the performance of transformer models compared to \ours with pure global attention. Surprisingly for dishwasher, pure local attention results in the best performance with MAE improved by 23.3\% and MCC improved by 28.9\%. Fridge demonstrates similar trends, while the best performance is achieved with 2 local attention heads. The results suggest that inductive bias in local context is helpful to improve energy disaggregation performance.

\subsubsection{Local Window Size}
\begin{table}[h]
  \centering
  \caption[Sensitivity of local window size]{Evaluation performance with different number of local window sizes.}
  \begin{tabular}{ccc}
  \toprule
  \multirow{2}{*}{\textbf{Window size}} & \multicolumn{1}{c}{\textbf{Microwave}} & \textbf{Washer} \\
  & \textbf{MAE} $\downarrow$ / \textbf{F1} $\uparrow$ / \textbf{MCC} $\uparrow$ & \textbf{MAE} $\downarrow$ / \textbf{F1} $\uparrow$ / \textbf{MCC} $\uparrow$ \\ \midrule
  $l_{\mathrm{win}} = 10$  & 18.35 / 0.29 / 0.29                                  & 40.83 / 0.34 / 0.33                                  \\
  $l_{\mathrm{win}} = 15$  & \underline{16.69} / \underline{0.30} / 0.34          & 41.40 / 0.36 / 0.35                                  \\
  $l_{\mathrm{win}} = 20$  & \textbf{15.82} / 0.29 / \underline{0.34}             & \textbf{39.35} / \underline{0.40} / \underline{0.39} \\
  $l_{\mathrm{win}} = 25$  & 17.95 / 0.28 / 0.34                                  & \underline{39.78} / \textbf{0.40} / \textbf{0.41}    \\
  $l_{\mathrm{win}} = 30$  & 17.42 / \textbf{0.31} / \textbf{0.41}                & 40.66 / 0.38 / 0.34                                  \\ \bottomrule
  \end{tabular}
  \label{tab:local-window}
\end{table}
We evaluate the influence of the window size in local attention heads. We initiate \ours with different values ranging from 10 to 30 and train on microwave and washer in REDD. Results are shown in \Cref{tab:local-window}. Despite certain variations, we observe the best window size for local attention heads is between 20 and 30 for microwave and washer. Overall, \ours with fixed $l_{\mathrm{win}} = 20$ yields the best MAE results, but does not guarantee best F1 and MCC. This suggests that the best local window size can be different depending on the individual appliance and evaluation method.
\section{Conclusion}
\label{sec:conclusion}

This work contributes a novel transformer model for the seq2point setting in NILM. The proposed \ours is designed for efficient energy disaggregation in smart sensor systems with limited computing power. We leverage efficient linear self-attention in global attention to reduce the computational complexity in both space and time. Moreover, we address localness modeling with local attention heads and relative positional encodings to improve the capture of local patterns. With extensive experiments and considerable improvements to state-of-the-art approaches, we demonstrate both the effectiveness and efficiency of \ours for energy disaggregation in smart residential systems.


\section*{Acknowledgment}

This research is supported in part by the National Science Foundation under Grant IIS-2008228, CNS-1845639, CNS-1831669. The views and conclusions contained in this document are those of the authors and should not be interpreted as representing the official policies, either expressed or implied, of the U.S. Government. The U.S. Government is authorized to reproduce and distribute reprints for Government purposes notwithstanding any copyright notation here on.

\bibliographystyle{IEEEtran}
\bibliography{reference}

\end{document}